 \pdfoutput=1
 
\documentclass[11pt]{article}
\usepackage{url}

\usepackage[hyperref]{acl2021}
\usepackage{times}
\usepackage{latexsym}

\usepackage{amsthm,amsmath,amssymb}
\usepackage{mathrsfs}
\usepackage{graphicx}
\usepackage{booktabs}
\usepackage{ulem}
\usepackage{multirow}

\usepackage{microtype}

\aclfinalcopy 
\def\aclpaperid{958} 

\title{A Sequence-to-Sequence Approach to Dialogue State Tracking}

\author{Yue Feng$^\dagger$ \thanks{The work was done when the first author was an intern at ByteDance AI Lab.} \quad Yang Wang$^\ddagger$ \quad Hang Li$^\ddagger$\\
  $^\dagger$University College London, London, UK\\
  $^\ddagger$ByteDance AI Lab, Beijing, China\\
  $^\dagger$ {\texttt{yue.feng.20@ucl.ac.uk}} \\
  $^\ddagger$\texttt{\{wangyang.127, lihang.lh\}@bytedance.com} \\}

\date{}

\begin{document}
\maketitle
\begin{abstract}

This paper is concerned with dialogue state tracking (DST) in a task-oriented dialogue system. Building a DST module that is highly effective is still a challenging issue, although significant progresses have been made recently. This paper proposes a new approach to dialogue state tracking, referred to as Seq2Seq-DU, which formalizes DST as a sequence-to-sequence problem. Seq2Seq-DU employs two BERT-based encoders to respectively encode the utterances in the dialogue and the descriptions of schemas, an attender to calculate attentions between the utterance embeddings and the schema embeddings, and a decoder to generate pointers to represent the current state of dialogue. Seq2Seq-DU has the following advantages. It can jointly model intents, slots, and slot values; it can leverage the rich representations of utterances and schemas based on BERT; it can effectively deal with categorical and non-categorical slots, and unseen schemas. In addition, Seq2Seq-DU can also be used in the NLU (natural language understanding) module of a dialogue system. Experimental results on benchmark datasets in different settings (SGD, MultiWOZ2.2, MultiWOZ2.1, WOZ2.0, DSTC2, M2M, SNIPS, and ATIS) show that Seq2Seq-DU outperforms the existing methods.
\end{abstract}

\begin{figure}[!t]
\centering
\includegraphics[width=0.48\textwidth]{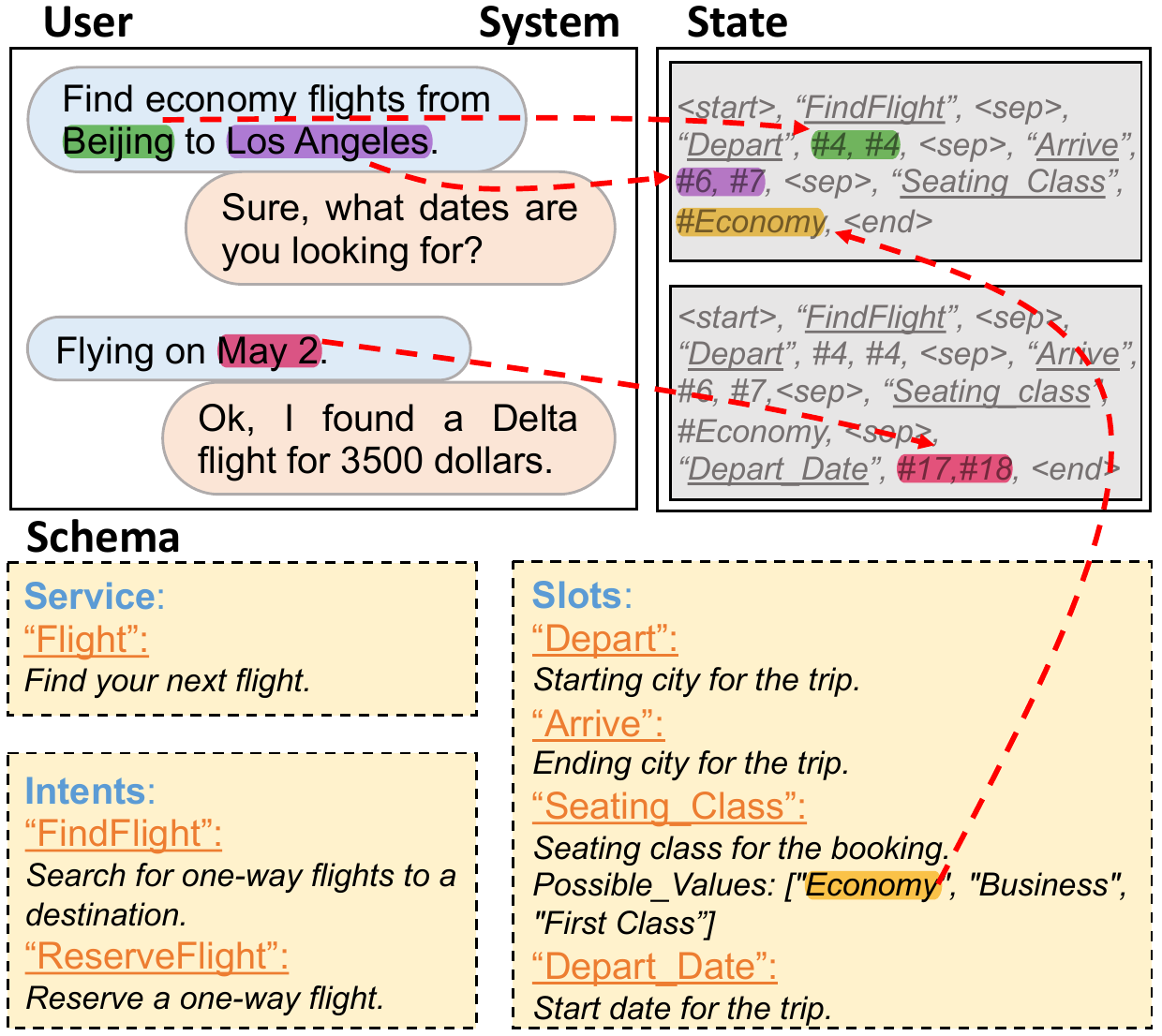}
\caption{An example of dialogue state tracking. Given a dialogue history that contains user utterances and system utterances, and descriptions of schema that contain all possible intents and slot-value pairs, a dialogue state for the current turn is created which is represented by intents and slot-value pairs. There are slot values obtained from the schema (categorical) as well as slot values extracted from the utterances (non-categorical). \#4, \#6, etc denote pointers.}
\label{fig:example}
\end{figure}

\section{Introduction}

A task-oriented dialogue system usually consists of several modules: natural language understanding (NLU), dialogue state tracking (DST), dialogue policy (Policy), and natural language generation (NLG). We consider DST and also NLU in this paper. In NLU, a semantic frame representing the content of user utterance is created in each turn of dialogue. In DST, several semantic frames representing the `states' of dialogue are created and updated in multiple turns of dialogue. Domain knowledge in dialogues is represented by a representation referred to as schema, which consists of possible intents, slots, and slot values. Slot values can be in a pre-defined set, with the corresponding slot being referred to as categorical slot, and they can also be from an open set, with the corresponding slot being referred to as non-categorical slot. Figure~\ref{fig:example} shows an example of DST.

We think that a DST module (and an NLU module) should have the following abilities. (1) Global, the model can jointly represent intents, slots, and slot values. (2) Represenable, it has strong capability to represent knowledge for the task, on top of a pre-trained language model like BERT. (3) Scalable, the model can deal with categorical and non-categorical slots and unseen schemas. 

Many methods have been proposed for DST ~\citep{wu2019transferable,zhong2018global,mrkvsic2016neural,goo2018slot}. There are two lines of relevant research. (1) To enhance the 
scalability of DST, a problem formulation, referred to as schema-guided dialogue, is proposed.  In the setting, it is assumed that descriptions on schemas in natural language across multiple domains are given and utilized. Consequently, a number of methods are developed to make use of schema descriptions to increase the scalability of DST~\cite{rastogi2019towards, zang2020multiwoz, noroozi2020fast}. The methods regard DST as a classification and/or an extraction problem and independently infer the intent and slot value pairs for the current turn. Therefore, the proposed models are generally representable and scalable, but not global. 
(2) There are also a few methods which view DST as a sequence to sequence problem. Some methods sequentially infer the intent and slot value pairs for the current turn on the basis of dialogue history and usually employ a hierarchical structure (not based on BERT) for the inference~\cite{lei2018sequicity, ren2019scalable, chen2020credit}. Recently, a new approach is proposed which formalizes the tasks in dialogue as sequence prediction problems using a unified language model (based on GPT-2)~\citep{hosseini2020simple}. The method cannot deal with unseen schemas and intents, however, and thus is not scalable.

We propose a novel approach to DST, referred to as Seq2Seq-DU (sequence-to-sequence for dialogue understanding), which combines the advantages of the existing approaches. To the best of our knowledge, there was no previous work which studied the approach. We think that DST should be formalized as a sequence to sequence or `translation' problem in which the utterances in the dialogue are transformed into semantic frames. In this way, the intents, slots, and slot values can be jointly modeled. Moreover, NLU can also be viewed as a special case of DST and thus Seq2Seq-DU can also be applied to NLU. We note that very recently the effectiveness of the sequence to sequence approach has also been verified in other language understanding tasks~\cite{paolini2021structured}.

Seq2Seq-DU comprises a BERT-based encoder to encode the utterances in the dialogue, a BERT based encoder to encode the schema descriptions, an attender to calculate attentions between the utterance embeddings and schema embeddings, and a decoder to generate pointers of items representing the intents and slots-value pairs of state.

Seq2Seq-DU has the following advantages. (1) Global: it relies on the sequence to sequence framework to simultaneously model the intents, slots, and slot-values.  (2) Representable: It employs BERT~\cite{devlin2018bert} to learn and utilize better representations of not only the current utterance but also the previous utterances in the dialogue. If schema descriptions are available, it also employs BERT for the learning and utilization of their representations. (3) Scalable: It uses the pointer generation mechanism, as in the Pointer Network~\cite{vinyals2015pointer}, to create representations of intents, slots, and slot-values, no matter whether the slots are categorical or non-categorical, and whether the schemas are unseen or not.

Experimental results on benchmark datasets show that Seq2Seq-DU\footnote{The code is available at \url{https://github.com/sweetalyssum/Seq2Seq-DU}.} performs much better than the baselines on SGD, MultiWOZ2.2, and MultiWOZ2.1 in multi-turn dialogue with schema descriptions, is superior to BERT-DST on WOZ2.0, DSTC2, and M2M, in multi-turn dialogue without schema descriptions, and works equally well as Joint BERT on ATIS and SNIPS in single turn dialogue (in fact, it degenerates to Joint BERT).

\begin{table*}[h]
\centering
\resizebox{1.0\textwidth}{!}{
\begin{tabular}{c|p{12cm}|c|c}
\toprule
\bf{Model} & \bf{Characteristics} & \bf{Data Sets} & \bf{Comparison} \\
\hline
\hline
FastSGD~\citep{noroozi2020fast}& BERT-based model, employs two carry-over procedures and multi-head attentions to model schema descriptions. & SGD & Yes \\
\hline
SGD Baseline~\citep{rastogi2019towards} & BERT-based model, predictions are made over a dynamic set of intents and slots, using their descriptions. & SGD and MultiWOZ2.2 & Yes \\
\hline
TripPy~\citep{heck2020trippy} & BERT-based model, make use of various copy mechanisms to fill slots with values. & MultiWOZ2.2 & Yes\\
\hline
TRADE~\citep{wu2019transferable} & Generate dialogue states from utterances using a copy mechanism, facilitating knowledge transfer for new schema elements. & MultiWOZ2.2 & Yes \\
\hline
DS-DST~\citep{zhang2019find} & BERT-based model, to classify over a candidate list or find values from text spans. & MultiWOZ2.2 & Yes \\
\hline
\hline
BERT-DST~\citep{chao2019bert} & Use BERT as dialogue context encoder and makes parameter sharing across slots. & DSTC2, WOZ2.0, and M2M& Yes \\
\hline
StateNet~\citep{ren2018towards} & Independent of number of values, shares parameters across slots and uses pre-trained word vectors. & DSTC2 and WOZ2.0& Yes \\
\hline
GLAD~\citep{zhong2018global} & Use global modules to share parameters across slots and uses local modules to retrain slot-specific parameters. & DSTC2 and WOZ2.0& Yes \\
\hline
Belief Tracking~\citep{ramadan2018large} & Utilize semantic similarity between dialogue utterances and ontology, and information is shared across domains. & DSTC2 and WOZ2.0& Yes \\
\hline 
Neural Belief Tracker~\citep{mrkvsic2016neural} & Conduct reasoning on pre-trained word vectors, and combines them into representations of user utterance and dialogue context. & DSTC2 and WOZ2.0& Yes \\
\hline
DST+LU~\citep{rastogi2018multi} & Select candidates for each slot, while candidates are generated by NLU. & M2M
 & Yes \\
\hline
\hline
Joint BERT~\citep{chen2019bert} & A joint intent classification and slot filling model based on BERT. & ATIS and SNIPS & Yes \\
\hline
Slot-Gated~\citep{goo2018slot} & Use a slot gate, models relation between intent and slot vectors to create semantic frames. & ATIS and SNIPS & Yes \\
\hline
Atten.-BiRNN~\citep{liu2016attention} & Attention-based model, explores several strategies for alignment between intent classification and slot labeling.  & ATIS and SNIPS & Yes \\
\hline
RNN-LSTM~\citep{hakkani2016multi} & Use RNN with LSTM cells to create complete semantic frames from user utterances. & ATIS and SNIPS & Yes \\
\hline
\hline
Sequicity~\cite{lei2018sequicity} & Two-stage sequence-to-sequence model based on CopyNet, conducts both dialogue state tracking and response generation. & CamRest676 and KVRET& No \\
\hline
COMER~\cite{ren2019scalable} &  BERT-based hierarchical encoder-decoder model, generates state sequence based on user utterance & WOZ2.0 and MultiWOZ2.0 & Yes \\
\hline
CREDIT~\cite{chen2020credit} & Hierarchical encoder-decoder model, views DST as a sequence generation problem.  & MultiWOZ2.0 and MultiWOZ2.1 & No \\
\hline
SimpleTOD~\cite{hosseini2020simple} & A unified sequence-to-sequence model based on GPT-2, conducts dialogue state tracking, dialogue action prediction, and response generation.  & MultiWOZ2.0 and MultiWOZ2.1 & Yes \\
\toprule
\end{tabular}
}
\caption{Summary of existing methods on DST.}
\label{tab:related_work}
\end{table*}

\section{Related Work}

There has been a large amount of work on task-oriented dialogue, especially dialogue state tracking and natural language understanding
(eg.,~\citep{zhang2020recent, huang2020challenges, chen2017survey}).
Table~\ref{tab:related_work} makes a summary of existing methods on DST. We also indicate the methods on which we make comparison in our experiments.

\subsection{Dialogue State Tracking}
Previous approaches mainly focus on encoding of the dialogue context and employ deep neural networks such as CNN, RNN, and LSTM-RNN to independently infer the values of slots in DST~\citep{mrkvsic2016neural, xu2018end, zhong2018global, ren2018towards, rastogi2017scalable, ramadan2018large, wu2019transferable, zhang2019find, heck2020trippy}. The approaches cannot deal with unseen schemas in new domains, however. To cope with the problem, a new direction called schema-guided dialogue is proposed recently, which assumes that natural language descriptions of schemas are provided and can be used to help transfer knowledge across domains. As such, a number of methods are developed in the recent dialogue competition SGD~\citep{rastogi2019towards, zang2020multiwoz, noroozi2020fast, chen2020schema}. 
Our work is partially motivated by the SGD initiative. Our model Seq2Seq-DU is unique in that it formalizes schema-guided DST as a sequence-to-sequence problem using BERT and pointer generation. 

In fact, sequence-to-sequence models are also utilized in DST. Sequicity~\citep{lei2018sequicity} is a two-step sequence to sequence model which first encodes the dialogue history and generates a belief span, and then generates a language response from the belief span. COMER~\citep{ren2019scalable} and CREDIT~\citep{chen2020credit} are hierarchical sequence-to-sequence models which represent the 
intents and slot-value pairs in a hierarchical way, and employ a multi-stage decoder. SimpleTOD~\citep{hosseini2020simple} is a unified approach to task-oriented dialogue which employs a single and causal language model to perform sequence prediction in DST, Policy, and NLG. Our proposed approach also uses a sequence-to-sequence model. There are significant differences between our model Seq2Seq-DU and the existing models. First, there is no hierarchy in decoding of Seq2Seq-DU. A flat structure on top of BERT appears to be sufficient for jointly capturing the intents, slots, and values. Second, the decoder in Seq2Seq-DU generates pointers instead of tokens, and thus can easily and effectively handle categorical slots, non-categorical slots, as well as unseen schemas.


\begin{figure*}[!t]
\centering
\includegraphics[width=0.92\textwidth]{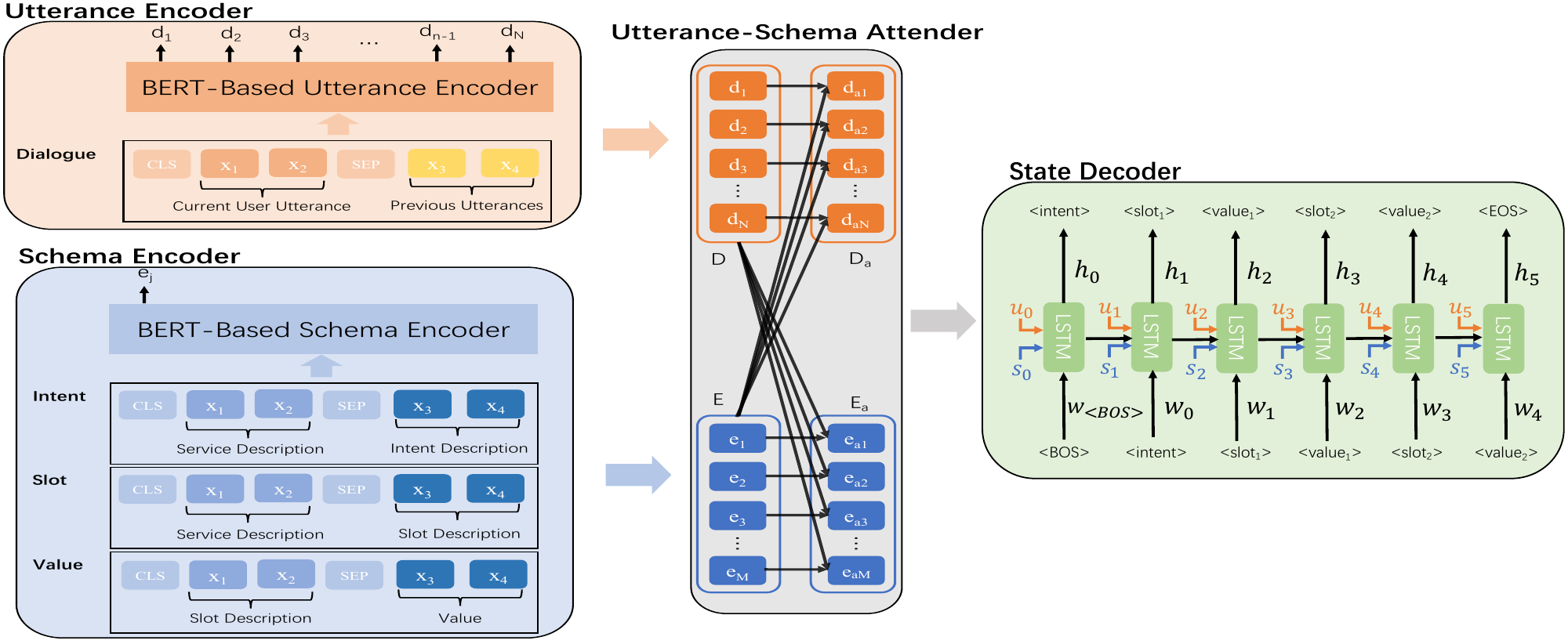}
\caption{The architecture of Seq2Seq-DU, containing utterance encoder, schema encoder, utterance-schema attender, and state decoder.}
\label{fig:encoder}
\end{figure*}

\subsection{Natural Language Understanding}
Traditionally the problem of NLU is decomposed into two independent issues, namely classification of intents and sequence labeling of slot-value pairs~\citep{liu2016attention, hakkani2016multi}. For example, deep neural network combined with conditional random field is employed for the task~\citep{yao2014recurrent}. Recently the pre-trained language model BERT~\citep{chen2019bert} is exploited to further enhance the accuracy. Methods are also proposed which can jointly train and utilize classification and sequence labeling models~\citep{chen2019bert, goo2018slot}. In this paper, we view NLU as special case of DST and employ our model Seq2Seq-DU to perform NLU. Seq2Seq-DU can degenerate to a BERT based NLU model.

\section{Our Approach}
\label{Approach}

Our approach Seq2Seq-DU formalizes dialogue state tracking as a sequence to sequence problem using BERT and pointer generation.
As shown in Figure~\ref{fig:encoder}, Seq2Seq-DU consists of an utterance encoder, a schema encoder, an utterance schema attender, and a state decoder. In each turn of dialogue, the utterance encoder transforms the current user utterance and the previous utterances in the dialogue into a sequence of utterance embeddings using BERT; the schema encoder transforms the schema descriptions into a set of schema embeddings also using BERT; the utterance schema attender calculates attentions between the utterance embeddings and the schema embeddings to create attended utterance and schema representations; finally, the state decoder sequentially generates a state representation on the basis of the attended representations using LSTM and pointer generation. 

\subsection{Utterance Encoder}
The utterance encoder takes the current user utterance as well as the previous utterances (user and system utterances) in the dialogue (a sequence of tokens) as input and employs BERT to construct a sequence of utterance embeddings. The relations between the current utterance and the previous utterances are captured by the encoder.

The input of the encoder is a sequence of tokens with length $N$, denoted as $X = (x_1, ..., x_N)$. The first token $x_1$ is [CLS], followed by the tokens of the current user utterance and the tokens of the previous utterances, separated by [SEP]. The output is a sequence of embeddings also with length $N$, denoted as $D = (d_1, ..., d_N)$ and referred to as utterance embeddings, with one embedding for each token. 

\subsection{Schema Encoder}
The schema encoder takes the descriptions of intents, slots, and categorical slot values (a set of combined sequences of tokens) as input and employs BERT to construct a set of schema embeddings.

\begin{table}[!h]
\centering
\resizebox{0.48\textwidth}{!}{
\begin{tabular}{c|c|c}
\toprule
\bf{Schema} & \bf{Sequence 1} & \bf{Sequence 2} \\
\hline
\hline
\textit{Intent}& service description & intent description \\
\hline
\textit{Slot}& service description & slot description \\
\hline
\textit{Value}& slot description & value \\
\toprule
\end{tabular}
}
\caption{Descriptions for a dialogue schema. Two combined descriptions are used for describing an intent, a slot, or a value in the schema.} 
\label{tab:schema_encoder}
\end{table}

Suppose that there are $I$ intents, $S$ slots, and $V$ categorical slot values in the schemas. Each schema element is described by two descriptions as outlined in Table~\ref{tab:schema_encoder}. The input is a set of combined sequences of tokens, denoted as $Y = \{y_1, ..., y_M\}$. Note that $M=I+S+V$. Each combined sequence starts with [CLS], followed by the tokens of the two descriptions with [SEP] as a separator. The final representation of [CLS] is used as the embedding of the input intent, slot, or slot value. The output is a set of embeddings, and all the embeddings are called schema embeddings $E = \{e_1, ..., e_{M}\}$. 

The schema encoder in fact adopts the same approach of schema encoding as in~\citep{rastogi2019towards}. There are two advantages with the approach. First, the encoder can be trained across different domains. Schema descriptions in different domains can be utilized together. Second, once the encoder is fine-tuned, it can be used to process unseen schemas with new intents, slots, and slot values.

\subsection{Utterance-Schema Attender}

The utterance-schema attender takes the sequence of utterance embeddings and the set of schema embeddings as input and calculates schema-attended utterance representations and utterance-attended schema representations. In this way, information from the utterances and information from the schemas are fused.

First, the attender constructs an attention matrix, indicating the similarities between utterance embeddings and schema embeddings. Given the $i$-th   utterance token embedding $d_i$ and $j$-th schema embedding $e_j$, it calculates the similarity as follows,
\begin{equation}\label{eq:overall}
A(i,j) = r^\intercal \text{tanh}(W_1 d_i + W_2 e_j),
\end{equation}
where $r$, $W_1$, $W_2$ are trainable parameters. 


The attender then normalizes each row of matrix $A$ as a probability distribution, to obtain matrix $\overline{A}$. Each row represents the attention weights of schema elements with respect to an utterance token. Then the schema-attended utterance representations are calculated as $D_a =  E \overline{A}^\intercal$. The attender also normalizes each column of matrix $A$ as a probability distribution, to obtain matrix $\widetilde{A}$. Each column represents the attention weights of utterance tokens with respect to a schema element. Then the utterance-attended schema representations are calculated as $E_a = D \widetilde{A}$.


\subsection{State Decoder}

The state decoder sequentially generates a state representation (semantic frame) for the current turn, which is represented as a sequence of pointers to elements of the schemas and tokens of the utterances (cf., Figure~\ref{fig:example}). The sequence can then be either re-formalized as a semantic frame in dialogue state tracking\footnote{For simplicity, we assume here that there is only one semantic frame in each turn. In principle, there can be multiple frames.},
\[ [intent; (slot_1, value_1); (slot_2, value_2); ...],\]
or a sequence of labels in NLU (intent-labeling and slot-filling). The pointers point to the elements of intents, slots, and slot values in the schema descriptions (categorical slot values), as well as the tokens in the utterances (non-categorical slot values). The elements in the schemas can be either words or phrases, and the tokens in the utterances form spans for extraction of slot values. 

The state decoder is an LSTM using pointer~\citep{vinyals2015pointer} and attention~\citep{bahdanau2014neural}. It takes the two representations $D_a$ and $E_a$ as input.
At each decode step $t$, the decoder receives the embedding of the previous item $w_{t-1}$, the utterance context vector $u_{t}$, the schema context vector $s_{t}$, and the previous hidden state $h_{t-1}$, and produces the current hidden state $h_t$:
\begin{equation}
h_t = \text{LSTM}(w_{t-1}, h_{t-1}, u_{t}, s_{t}).
\end{equation}

We adopt the attention function in~\citep{bahdanau2014neural} to calculate the context vectors as follows,
\begin{gather}
u_{t} = \text{attend}(h_{t-1}, D_a, D_a),\\
s_{t} = \text{attend}(h_{t-1}, E_a, E_a).
\end{gather}

The decoder then generates a pointer from the set of pointers in the schema elements and the tokens of the utterances on the basis of the hidden state $h_{t}$.
Specifically, it generates a pointer of item $w$ according to the following distribution,
\begin{gather}
z_w = q^\intercal \text{tanh}(U_1 h_t + U_2 k_w), \\
P(\# w) = \text{softmax}(z_w),
\end{gather}
where $\# w$ is the pointer of item $w$, $k_w$ is the representation of item $w$ either in the utterance representations $D_a$ or in the schema representations $E_a$, $q$, $U_1$, and $U_2$ are trainable parameters, and softmax is calculated over all possible pointers.

During decoding, the decoder employs beam search to find the best sequences of pointers in terms of probability of sequence.

\begin{table*}[h]
\centering
\resizebox{1.0\textwidth}{!}{
\begin{tabular}{l|cccccccc}
        \toprule
        { \bf{Characteristics}}&{ \bf{SGD}}&{ \bf{MultiWOZ2.2}} & { \bf{MultiWOZ2.1}} &{ \bf{WOZ2.0}} &{ \bf{DSTC2}} &{ \bf{M2M}} &{ \bf{ATIS}} &{ \bf{SNIPS}}\\
 		\hline
        \hline
        \text{No. of domains} & 16 & 8 & 7& 1 & 1 & 2 &- &-\\
        \text{No. of dialogues} & 16,142 & 8,438 & 8438& 1,612 & 600 & 1,500 & 4,478 & 13,084\\
        \text{Total no. of turns} & 329,964 & 113,556& 113,556& 23,354 & 4,472 & 14,796 & 4,478 & 13,084\\
        \text{Avg. turns per dialogue} & 20.44 & 13.46&13.46& 14.49 & 7.45 & 9.86 & 1 & 1\\
        \text{Avg. tokens per turn} & 9.75 & 13.13&13.38& 8.54 & 11.24 & 8.24 & 11.28 & 9.09\\
        \text{No. of categorical slots} & 53 & 21 &37& 3 & 3 & 0 & 0 & 0\\
        \text{No. of non-categorical slots} & 162 & 40 &0& 0 & 0 & 14 & 120 & 72\\
		\text{Have schema description} & Yes & Yes &Yes& No & No & No & No & No\\
		\text{Have unseen schemas in test set} & Yes &No &No& No & No & No & No & No\\
		\toprule
	\end{tabular}
}
\caption{Statistics of datasets in experiments. Numbers are those of training datasets.}
\label{tab:datasets}
\end{table*}

\subsection{Training}

The training of Seq2Seq-DU follows the standard procedure of sequence-to-sequence. The only difference is that it is always conditioned on the schema descriptions.  Each instance in training consists of the current utterance and the previous utterances, and the state representation (sequence of pointers) for the current turn.  Two pre-trained BERT models are used for representations of utterances and schema descriptions respectively. The BERT models are then fine-tuned in the training process. Cross-entropy loss is utilized to measure the loss of generating a sequence.

\section{Experiments}

\subsection{Datasets}
We conduct experiments using the benchmark datasets on task-oriented dialogue. SGD~\citep{rastogi2019towards} and MultiWOZ2.2~\citep{zang2020multiwoz} are datasets for DST; they include schemas with categorical slots and non-categorical slots in multiple domains and natural language descriptions on the schemas, as shown in Table~\ref{tab:schema_encoder}. In particular, SGD includes unseen schemas in the test set. 
MultiWOZ2.1~\citep{eric2020multiwoz} is the previous version of MultiWOZ2.2, which only has categorical slots in multiple domains.
WOZ2.0~\citep{wen2016network} and DSTC2~\citep{henderson2014second} are datasets for DST; they contain schemas with only categorical slots in a single domain. M2M~\citep{shah2018building} is a dataset for DST and it has span annotations for slot values in multiple domains. ATIS~\citep{tur2010left} and SNIPS~\citep{coucke2018snips} are datasets for NLU in single-turn dialogues in a single domain. Table~\ref{tab:datasets} gives the statics of datasets in the experiments.


\subsection{Baselines and Variants}
We make comparison between our approach and the state-of-the-art methods on the datasets.


\noindent\textbf{SGD, MultiWOZ2.2 and MultiWOZ2.1}:
We compare Seq2SeqDU with six state-of-the-art methods on SGD, MultiWOZ2.2 and MultiWOZ2.1, which utilize schema descriptions, span-based and candidate-based methods, unified seq2seq model and BERT: FastSGT~\citep{noroozi2020fast}, SGDbaseline~\citep{rastogi2019towards}, TripPy~\citep{heck2020trippy}, SimpleTOD~\citep{hosseini2020simple}, TRADE~\citep{wu2019transferable}, and DS-DST~\citep{zhang2019find}.

\noindent\textbf{WOZ2.0 and DSTC2}: Our approach is compared against the state-of-the-art methods on WOZ2.0 and DSTC2, including those using a hierarchical seq2seq model and BERT: COMER~\cite{ren2019scalable}, BERT-DST~\citep{chao2019bert}, StateNet~\citep{ren2018towards}, GLAD~\citep{zhong2018global}, Belief Tracking~\citep{ramadan2018large}, and Neural Belief Tracker~\citep{mrkvsic2016neural}.

\noindent\textbf{M2M}:
We evaluate our approach and the state-of-the-art methods on M2M, which respectively employ a BERT-based architecture and a jointly-trained language understanding model, BERT-DST~\citep{chao2019bert} and DST+LU~\citep{rastogi2018multi}.

\noindent\textbf{ATIS and SNIPS}:
We make comparison between our approach and the state-of-the-art methods on ATIS and SNIPS for NLU within the sequence labeling framework, including Joint BERT~\citep{chen2019bert}, Slot-Gated~\citep{goo2018slot}, Atten.-BiRNN~\citep{liu2016attention}, and RNN-LSTM~\citep{hakkani2016multi}.

We also include two variants of Seq2Seq-DU. The differences are whether to use the schema descriptions, and the formation of dialogue state.





\noindent\textbf{Seq2Seq-DU-w/oSchema}:
It is used for datasets that do not have schema descriptions. It only contains utterance encoder and state decoder.

\noindent\textbf{Seq2Seq-DU-SeqLabel}:
It is used for NLU in a single-turn dialogue. It views the problem as sequence labeling, and only contains the utterance encoder and state decoder.

\subsection{Evaluation Measures}
We make use of the following metrics in evaluation.

\noindent\textbf{Intent Accuracy}: percentage of turns in dialogue for which the intent is correctly identified.


\noindent\textbf{Joint Goal Accuracy}: 
percentage of turns for which all the slots are correctly identified. For non-categorical slots, a fuzzy matching score is used on SGD and exact match are used on the other datasets to keep the numbers comparable with other works.

\noindent\textbf{Slot F1}: 
F1 score to evaluate accuracy of slot sequence labeling.

\subsection{Training}
We use the pre-trained BERT model (\text{[BERT-Base, Uncased]}), which has 12 hidden layers of 768 units and 12 self-attention heads to encode utterances and schema descriptions. The hidden size of LSTM decoder is also 768. The dropout probability is 0.1. We also use beam search for decoding, with a beam size of 5. The batch size is set to 8. Adam~\citep{kingma2014adam} is used for optimization with an initial learning rate of 1e-4. Hyper parameters are chosen using the validation dataset in all cases.

The training curves of all models are shown in Appendix~\ref{sec:training_curves}.

\subsection{Experimental Results}
Tables~\ref{tab:SGD_MultiWOZ2.2},~\ref{tab:WOZ2.0_DSTC2}, ~\ref{tab:M2M}, and ~\ref{tab:ATIS_SNIPS} show the results. One can see that Seq2Seq-DU performs significantly better than the baselines in DST and performs equally well as the baselines in NLU.

DST is carried out in different settings in SGD, MultiWOZ2.2, MultiWOZ2.1, WOZ2.0, DSTC2, and M2M. In all cases, Seq2Seq-DU works significantly better than the baselines. The results indicate that Seq2Seq-DU is really a general and effective model for DST, which can be applied to multiple settings. Specifically, Seq2Seq-DU can leverage the schema descriptions for DST when they are available (SGD and MultiWOZ2.2, MultiWOZ2.1)\footnote{There are better performing systems in the SGD competition. The systems are not based on single methods and thus are not directly comparable with our method.}. It can work well in zero-shot learning to deal with unseen schemas (SGD). It can also effectively handle categorical slots (MultiWOZ2.1, WOZ2.0 and DSTC2) and non-categorical slots (M2M). 
It appears that the success of Seq2Seq-DU is due to its suitable architecture design with a sequence-to-sequence framework, BERT-based encoders, utterance-schema attender, and pointer generation decoder.

NLU is formalized as sequence labeling in ATIS and SNIPS. Seq2Seq-DU is degenerated to Seq2Seq-DU-SeqLabel, which is equivalent to the baseline of Joint Bert. The results suggest that it is the case. Specially, the performances of Seq2Seq-DU are comparable with Joint BERT, indicating that Seq2Seq-DU can also be employed in NLU.


\begin{table}[!h]
\centering
\resizebox{0.5\textwidth}{!}{
\begin{tabular}{l|cc|cc|cc}
        \toprule
        \multirow{2}{*}{\bf{Model}}&\multicolumn{2}{c|}{{ \bf{SGD}}}&\multicolumn{2}{c|}{{ \bf{MultiWOZ2.2}}}&\multicolumn{2}{c}{{ \bf{MultiWOZ2.1}}}\\
        \cline{2-7}
        &{Joint GA} &{Int Acc} &{Joint GA} &{Int Acc}&{Joint GA} &{Int Acc} \\
 		\hline
        \hline
        \text{SGD-baseline} & 0.254&  0.906& 0.420& - & 0.434 & -\\
        \text{TRADE} & -& -& 0.454& - & 0.460 & -\\
        \text{DS-DST} & -& -& 0.517& - & 0.512 & -\\
        \text{FastSGT} & 0.292& 0.903& -& - &- &-\\
        \text{SimpleTOD} & -& -& -& -& 0.514 & -\\
        \text{TripPy} & -& -& 0.535& - & 0.553 & -\\
        \hline
        \text{Seq2Seq-DU} & ${\bf 0.301}$ & ${\bf 0.910}$& ${\bf 0.544}$& ${\bf 0.909}$ & ${\bf 0.561}$ & ${\bf 0.911}$\\
		\toprule
	\end{tabular}
}
\caption{Accuracies of Seq2Seq-DU and baselines on SGD, MultiWOZ2.2 and MultiWOZ2.1 datasets. Seq2Seq-DU outperforms baselines in terms of all metrics.}
\label{tab:SGD_MultiWOZ2.2}
\end{table}

\begin{table}[!h]
\centering
\resizebox{0.4\textwidth}{!}{
\begin{tabular}{l|c|c}
        \toprule
        \multirow{2}{*}{{ \bf{Model}}}&\multicolumn{1}{c|}{{ \bf{WOZ2.0}}}&\multicolumn{1}{c}{{ \bf{DSTC2}}}\\
        \cline{2-3}
        &{Joint GA} &{Joint GA}  \\
 		\hline
        \hline
        \text{Neural Belief Tracker} & 0.842& 0.734\\
        \text{Belief Tracking} & 0.855& -\\
        \text{GLAD} & 0.881& 0.745\\
        \text{StateNet} & 0.889& 0.755\\
        \text{BERT-DST} & 0.877& 0.693\\
        \text{COMER} & 0.886& -\\
        \hline
		\text{Seq2Seq-DU-w/oSchema} & ${\bf 0.912}$ & ${ \bf 0.850}$ \\
		\toprule
	\end{tabular}
}
\caption{Accuracies of Seq2Seq-DU and baselines on WOZ2.0 and DSTC2 datasets. Seq2Seq-DU-w/oSchema performs significantly better than the baselines.}  
\label{tab:WOZ2.0_DSTC2}
\end{table}

\begin{table}[!h]
\centering
\resizebox{0.4\textwidth}{!}{
\begin{tabular}{l|cc}
        \toprule
        \multirow{2}{*}{{ \bf{Model}}}&\multicolumn{2}{c}{{ \bf{M2M}}}\\
        \cline{2-3}
        &{Joint GA} &{Int Acc}  \\
 		\hline
        \hline
        \text{DST+LU} & 0.767& -\\
        \text{BERT-DST} & 0.869& -\\
        \hline
		\text{Seq2Seq-DU-w/oSchema} & ${\bf 0.909}$ & ${ \bf 0.997}$ \\
		\toprule
	\end{tabular}
}
\caption{Accuracies of Seq2Seq-DU and baselines on M2M dataset. Seq2Seq-DU-w/oSchema significantly outperforms the baselines.}  
\label{tab:M2M}
\end{table}

\begin{table}[!h]
\centering
\resizebox{0.5\textwidth}{!}{
\begin{tabular}{l|cc|cc}
        \toprule
        \multirow{2}{*}{{ \bf{Model}}}&\multicolumn{2}{c|}{{ \bf{ATIS}}}&\multicolumn{2}{c}{{ \bf{SNIPS}}}\\
        \cline{2-5}
        &{Slot F1} &{Int Acc} &{Slot F1} &{Int Acc} \\
 		\hline
        \hline
        \text{RNN-LSTM} & 0.943& 0.926&0.873&0.969\\
        \text{Atten.-BiRNN} & 0.942& 0.911&0.878&0.967\\
        \text{Slot-Gated} & 0.952& 0.941&0.888&0.970\\
        \text{Joint BERT} & {\bf 0.961}& {\bf 0.975}&{\bf0.970}&0.986\\
        \hline
		\text{Seq2Seq-DU-SeqLabel} & $ 0.955$ & $ 0.968$ &0.965& {\bf 0.990}\\
		\toprule
	\end{tabular}
}
\caption{Accuracies of Seq2Seq-DU and baselines on ATIS and SNIPS datasets. Seq2Seq-DU-SeqLabel performs comparably with Joint BERT.}  
\label{tab:ATIS_SNIPS}
\end{table}

\subsection{Ablation Study}
We also conduct ablation study on Seq2Seq-DU. We validate the effects of three factors: BERT-based encoder, utterance-schema attention, and pointer generation decoder. The results indicate that all the components of Seq2Seq-DU are indispensable.

\subsubsection*{Effect of BERT}
To investigate the effectiveness of using BERT in the utterance encoder and schema encoder, we replace BERT with Bi-directional LSTM and run the model on SGD and MultiWOZ2.2. As shown in Figure~\ref{fig:attention}, the performance of the BiLSTM-based model Seq2Seq-DU-w/oBert in terms of Joint GA and Int. Acc decreases significantly compared with Seq2Seq-DU. It indicates that the BERT-based encoders can create and utilize more accurate representations for dialogue understanding.

\subsubsection*{Effect of Attention}
To investigate the effectiveness of using attention, we compare Seq2Seq-DU with Seq2Seq-DU-w/oAttention which eliminates the attention mechanism, Seq2Seq-DU-w/SchemaAtt which only contains the utterance-attended schema representations, and Seq2Seq-DU-w/UtteranceAtt which only contains the schema-attended utterance representations. Figure~\ref{fig:attention} shows the results on SGD and MultiWOZ2.2 in terms of Joint GA and Int. Acc.  One can observe that without attention the performances deteriorate considerably. In addition, the performances of unidirectional attentions are inferior to the performance of bidirectional attention. Thus, utilization of bidirectional attention between utterances and schema descriptions is desriable. 



\subsubsection*{Effect of Pointer Generation}
\begin{figure}[t]
\centering
\includegraphics[width=0.48\textwidth]{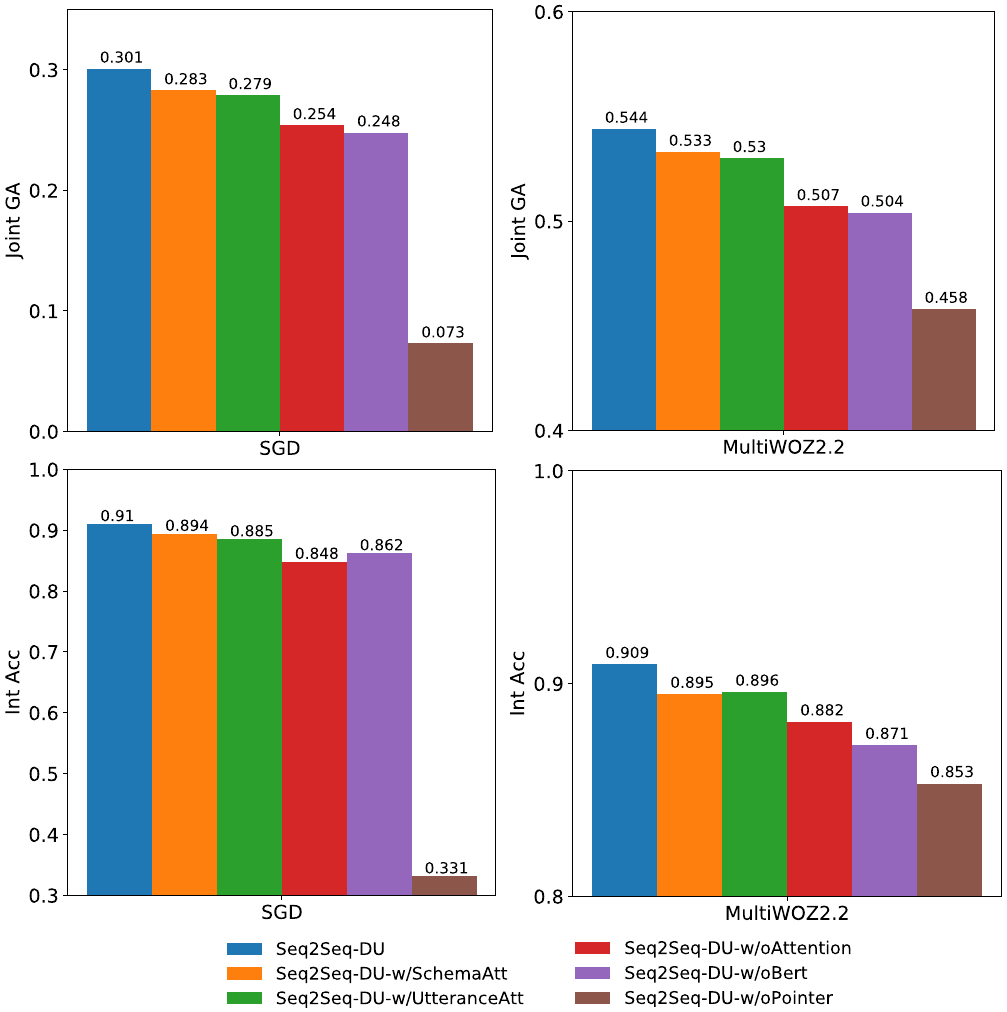}
\caption{Ablation study results of Seq2Seq-DU with respect to BERT, attention, and pointer generation on SGD and MultiWOZ2.2. }
\label{fig:attention}
\end{figure}

To investigate the effectiveness of the pointer generation mechanism, we directly generate words from the vocabulary instead of generating pointers in the decoding process. Figure~\ref{fig:attention} also shows the results of Seq2Seq-DU-w/oPointer on SGD and MultiWOZ2.2 in terms of Joint GA and Int. Acc. From the results we can see that pointer generation is crucial for coping with unseen schemas. In SGD which contains a large number of unseen schemas in the test set, there is significant performance degradation without pointer generation. The results on MultiWOZ2.2, which does not have unseen schemas in the test set, show pointer generation can also 
make significant improvement on already seen schemas by making full use of schema descriptions.

\begin{table*}[!h]
\centering
\resizebox{1.0\textwidth}{!}{
\begin{tabular}{|l|p{5cm}|p{4cm}|p{4cm}|p{4cm}|}
        \toprule
        \bf{ID}&\bf{Dialogue Utterance}&\bf{Dialogue State}&\bf{State Predictions of SGD-baseline}&\bf{State Predictions of Seq2Seq-DU}\\
 		\hline
        \hline
        \textit{1}  &
        \textit{User}: I wanna rent a place in Campbell. \textit{Sys}: How many baths? \textit{User}: One bath is fine. \textit{Sys}: How many bedrooms? \textit{User}: One bedroom is fine. It also needs in-unit laundry.&
        ``area": Campbell;
        ``\uline{in-unit-laundry": True};
        ``\uline{intent": rent};
        ``number-of-baths": 1;
        ``number-of-beds": 1;
        ``active-intent": FindHomeByArea;&
        ``area": Campbell;
        {\color{red} ``in-unit-laundry":  --  };
        {\color{blue} ``intent": rent};
        ``number-of-baths": 1;
        ``number-of-beds": 1; 
        ``active-intent": FindHomeByArea;& 
        ``area": Campbell;
        { \color{blue} ``in-unit-laundry": True};
        { \color{blue} ``intent": rent};
        ``number-of-baths": 1;
        ``number-of-beds": 1; 
        ``active-intent": FindHomeByArea;\\
        \hline
        \textit{2} & \textit{User}: The location isn't really important. It does need to be cheap though, and preferably a guesthouse.& 
        {``hotel-area": dontcare };
        ``\uline{hotel-pricerange": cheap};
        ``\uline{ hotel-type": guesthouse};
        ``active intent": find-hotel;& 
        {``hotel-area": dontcare };
        {\color{blue} ``hotel-pricerange": cheap};
        {\color{red} ``hotel-type": hotel};
        ``active intent": find-hotel;& 
        {``hotel-area": dontcare};
        {\color{blue}``hotel-pricerange": cheap};
        {\color{blue} ``hotel-type": guesthouse};
        ``active intent": find-hotel;\\
		\toprule
	\end{tabular}
}
\caption{Case study on Seq2Seq-DU and SGD-baseline on SGD and MultiWOZ2.2. The first example is from SGD and the second is from MultiWOZ2.2. The underlined slot-value pairs represent the ground truth states. The slot-value pairs in blue are correctly predicted ones, while the slot-value pairs in red are incorrectly predicted ones.}  
\label{tab:error}
\end{table*}

\subsection{Discussions}

\subsubsection*{Case Study}

We make qualitative analysis on the results of Seq2Seq-DU and SGD-baseline on SGD and MultiWOZ2.2. We find that Seq2Seq-DU can make more accurate inference of dialogue states by leveraging the relations existing in the utterances and schema descriptions.
For example, in the first case in Table~\ref{tab:error}, the user wants to find a cheap guesthouse. Seq2Seq-DU can correctly infer that the hotel type is ``guesthouse” by referring to the relation between ``hotel-pricerange" and ``hotel-type". In the second case, the user wants to rent a room with in-unit laundry. In the dataset, a user who intends to rent a room will care more about the laundry property. Seq2Seq-DU can effectively extract the relation between ``intent" and ``in-unit-laundry", yielding a correct result. In contrast, SGD-baseline does not model the relations in the schemas, and thus it cannot properly infer the values of ``hotel-type" and ``in-unit-laundry". 


\subsubsection*{Dealing with Unseen Schemas}

We analyze the zero-shot learning ability of Seq2Seq-DU.
Table~\ref{tab:domain_performance_sgd} presents the accuracies of Seq2Seq-DU in different domains on SGD. (Note that only SGD has unseen schemas in test set.) We observe that the best performances can be obtained in the domains with all seen schemas. The domains that have more partially seen schemas achieve higher accuracies, such as "Hotels", "Movies", "Services". The accuracies decline in the domains with more unseen schemas, such as "Messaging" and "RentalCars". We conclude that Seq2Seq-DU can perform zero-shot learning across domains. However, the ability still needs enhancement.

\begin{table}[!h]
\centering
\resizebox{0.5\textwidth}{!}{
\begin{tabular}{l|c|c||c|c|c}
        \toprule
        \textbf{Domain} & \textbf{Joint GA}& \textbf{Int Acc} & \textbf{Domain} & \textbf{Joint GA} & \textbf{Int Acc}\\
        \hline
        \textit{Messaging*} & 0.0489 & 0.3510 & \textit{Media*} & 0.2307 & 0.9065 \\
        \textit{RentalCars*} & 0.0625 & 0.7901 & \textit{Events*} & 0.3186 & 0.9327\\
        \textit{Payment*} & 0.0719 & 0.5835 & \textit{Hotels**} & 0.3396 & 0.9891 \\
        \textit{Music*} & 0.1234 & 0.9438 & \textit{Movies**} & 0.4386 & 0.7836 \\
		\textit{Restaurants*} & 0.1295 & 0.9627 & \textit{Travel} & 0.4490 & 0.9966 \\
		\textit{Flights*} & 0.1589 & 0.9649 & \textit{Services**} & 0.4774 & 0.9842\\
		\textit{Trains*} & 0.1683 & 0.9257 & \textit{Alarm*} & 0.5567 & 0.5768 \\
		\textit{Buses*} & 0.1684 & 0.8805  & \textit{Weather} & 0.5792 & 0.9965 \\
		\textit{Homes} & 0.2275 & 0.9081 & \textit{RideSharing} & 0.6702 & 0.9991 \\
		\toprule
	\end{tabular}
}
\caption{Accuracy of Seq2Seq-DU in each domain on SGD test set. Domains marked with ‘*’ are those for which the schemas in the test set are not present in the training set. Domains marked with ‘**’ have both the unseen and seen schemas. For other domains, the schemas in the test set are also seen in the training set.}  
\label{tab:domain_performance_sgd}
\end{table}


\subsubsection*{Categorical Slots and Non-categorical Slots}

Table~\ref{tab:SGD_MultiWOZ2.2_v2} shows the accuracies of Seq2Seq-DU and the baselines with respect to categorical and non-categorical slots on SGD and MultiWOZ2.2. (We did not compare with FastSGT on SGD dataset due to unavailability of the codes.) One can see that Seq2Seq-DU can effectively deal with both categorical and non-categorical slots. Furthermore, Seq2Seq-DU demonstrates higher accuracies on categorical slots than non-categorical slots. We conjecture that it is due to the co-occurrences of categorical slot values in both the dialogue history and the schema descriptions. The utterance-schema attention can more easily capture the relations between the values.

\begin{table}[!h]
\centering
\resizebox{0.5\textwidth}{!}{
\begin{tabular}{l|p{2.2cm}p{2.3cm}|p{2.2cm}p{2.3cm}}
        \toprule
        \multirow{2}{*}{{ \bf{Model}}}&\multicolumn{2}{c|}{{ \bf{SGD}}}&\multicolumn{2}{c}{{ \bf{MultiWOZ2.2}}}\\
        \cline{2-5}
        &{Categorical-Joint-GA} &{Noncategorical-Joint-GA} &{Categorical-Joint-GA} &{Noncategorical-Joint-GA} \\
 		\hline
        \hline
        \text{TRADE} & -& -& 0.628& 0.666\\
        \text{SGD-baseline} & 0.513&  0.361& 0.570& 0.661\\
        \text{DS-DST} & -& -& 0.706& 0.701\\
        \text{FastSGT} & not available  & not available & -& -\\
        \text{TripPy} & - & - & 0.684& {\bf 0.733}\\
        \hline
		\text{Seq2Seq-DU} & ${\bf 0.578}$ & ${\bf 0.393}$& ${\bf 0.758}$& $0.711$\\
		\toprule
	\end{tabular}
}
\caption{Accuracies of Seq2Seq-DU and baselines with respect to categorical and non-categorical slots on SGD and MultiWOZ2.2.}  
\label{tab:SGD_MultiWOZ2.2_v2}
\end{table}

\section{Conclusion}
We have proposed a new approach to dialogue state tracking. The approach, referred to as Seq2Seq-DU, takes dialogue state tracking (DST) as a problem of transforming all the utterances in a dialogue into semantic frames (state representations) on the basis of schema descriptions. 
Seq2Seq-DU is unique in that within the sequence to sequence framework it employs  BERT in encoding of utterances and schema descriptions respectively and generates pointers in decoding of dialogue state. Seq2Seq-DU is a global, reprentable, and scalable model for DST as well as NLU (natural language understanding).  Experimental results show that Seq2Seq-DU significantly outperforms the state-of-the-arts methods in DST on the benchmark datasets of SGD, MultiWOZ2.2, MultiWOZ2.1, WOZ2.0, DSTC2, M2M, and performs as well as the state-of-the-arts in NLU on the benchmark datasets of ATIS and SNIPS. 


\bibliography{acl2021}
\bibliographystyle{acl_natbib}


\clearpage

\appendix

\section{Training Curves}
\label{sec:training_curves}

\begin{figure}[h]
\centering
\includegraphics[width=0.48\textwidth]{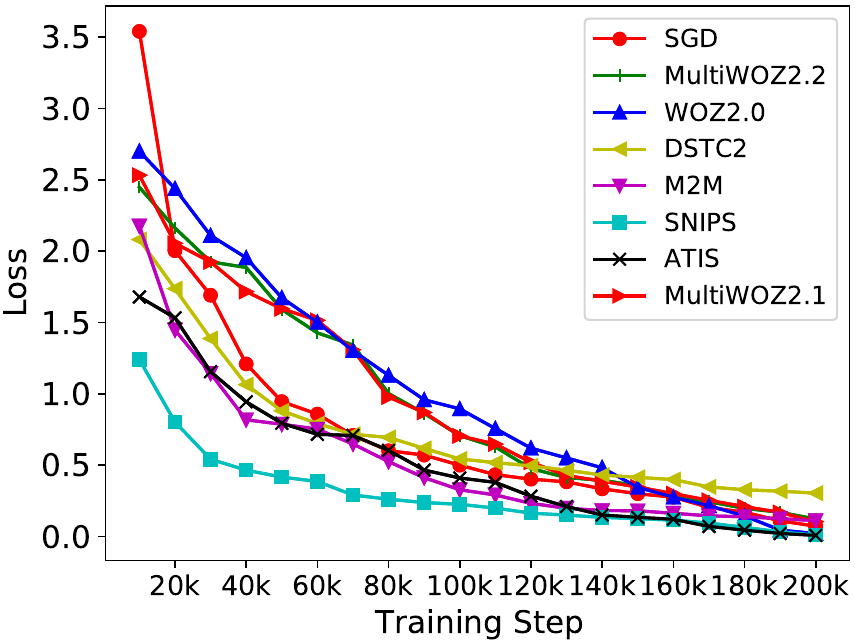}
\caption{Training losses of Seq2Seq-DU on all training datasets.}
\label{fig:loss}
\end{figure}

\begin{figure}[h]
\centering
\includegraphics[width=0.48\textwidth]{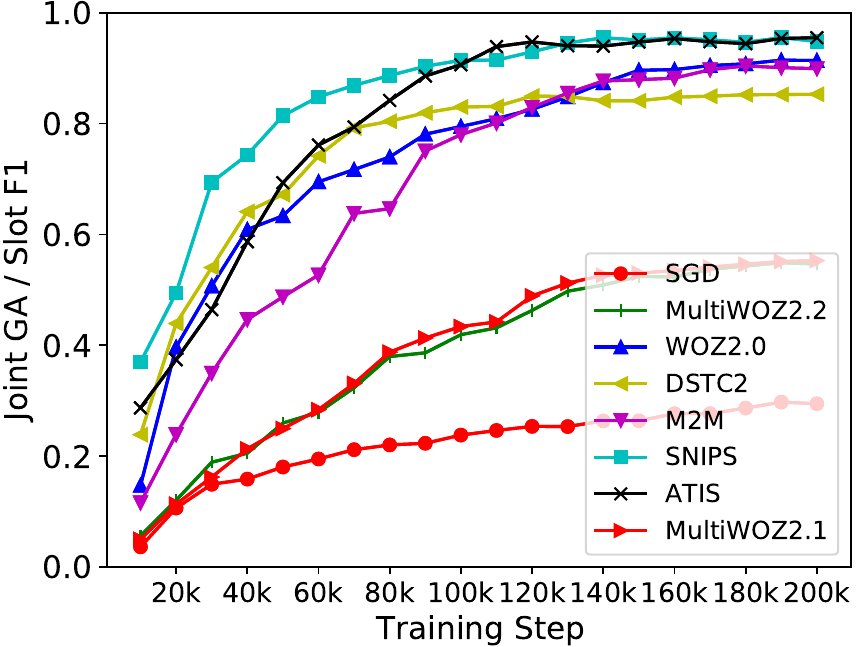}
\caption{Accuracies of Seq2Seq-DU in terms of Join GA / Slot F1 on all test sets.}
\label{fig:performance}
\end{figure}

Figure~\ref{fig:loss} shows the training losses of Seq2Seq-DU on the training datasets, while Figure~\ref{fig:performance} shows the accuracies of Seq2Seq-DU on the test sets during training. We regard training convergence when the fluctuation of loss is less than 0.01 for consecutive 20 thousand steps. Seq2Seq-DU converges at the 180k-th step on SGD, MultiWOZ2.2, and MultiWOZ2.1. Seq2Seq-DU-w/oSchema converges at the 150k-th step on WOZ2.0 and at the 140k-th step on DSTC2, and M2M. Furthermore, Seq2Seq-DU-SeqLabel converges at the 130k-th step on ATIS and SNIPS. These are consistent with the general trends in machine learning that more complex models are more difficult to train. 

\end{document}